\documentclass[10pt,twocolumn]{ICCAS}
\usepackage{diagbox}
\usepackage{subcaption}
\usepackage{soul}
\usepackage{float}
\usepackage{amsmath}     
\usepackage{amssymb}     
\usepackage{dsfont}      
\usepackage{multirow}    
\usepackage[table]{xcolor} 
\usepackage{booktabs}    
\usepackage{makecell}    
\newcommand{\thinhline}{\noalign{\hrule height 0.01pt}} 
\graphicspath{{figures/}}

\begin{document}

\title{Comparative Study on Agility, Efficiency, and Impact\\Absorption of Bipedal Robots with Active Toes}

\author{Joong-Gil Kim$^{1}$, Wontae Ye$^{1}$, Geunwoo Cho$^{1}$, Seong-Ho Yun$^{2}$, Se-Hyoung Cho$^{3}$ and Yong-Jae Kim$^{1,3,*}$}

\affils{$^{1}$School of Electrical, Electronics and Communication Engineering, Korea University of Technology and Education, \\
Cheonan, 31253, Korea (giribboy97@koreatech.ac.kr; solnox99@koreatech.ac.kr; jgw58@koreatech.ac.kr; yongjae@koreatech.ac.kr){\small${}^{*}$ Corresponding author} \\
$^{2}$Artificial Intelligence and Robotics Institute, Korea Institute of Science and Technology, \\
Seoul, 02792, South Korea (dnstjdgh@gmail.com) \\
$^{3}$Robot Innovation Hub, WIRobotics Inc., \\
Cheonan, 31253, Korea (shcho@wirobotics.com)}


\abstract{
    Human legs exhibit high efficiency, agility, and impact absorption, with toes playing a crucial role in these
    capabilities. While many attempts have been made to implement human-like toes in robots, they have not fully
    replicated human characteristics nor rigorously validated their benefits. We propose a 14-DOF biped robot emulating
    human toes' lightweight, high-torque, robust nature. To quantitatively analyze the effectiveness of the active
    toes in terms of agility, efficiency, and impact absorption, we developed a high-fidelity simulation training
    environment that reflects actual actuators with coupled transmissions and accurate power consumption.
    To ensure a fair comparison between configurations with and without active toes, we designed a minimal RL reward
    function and applied an identical training procedure to both. The simulation results indicate that, at
    1.33~m/s walking, the toe-equipped robot reduced CoT by 17.5\% and heel-strike GRF by 5.0\% compared with the
    toe-ablation configuration. On the agility test, average and maximum path deviation decreased by 25.0\% and
    34.0\%, respectively.
}


\maketitle


\section{Introduction}
Human locomotion is an intricate and highly efficient process, with the foot playing a crucial role in agility, efficiency, and impact absorption [1][2]. The human foot weighs approximately 1 kg and is composed of a 2-degree-of-freedom (DOF) ankle and multiple DOF toes [1]. Despite its relatively light weight, the foot can deliver up to 104 Nm of isometric torque and withstand ground reaction forces equivalent to 2.5 times body weight during running [1]-[3]. The human toe, in particular, contributes significantly to
stride length, stability, and propulsion, allowing for dynamic
adaptation to uneven terrain [4].

Inspired by these capabilities, the field of bipedal robotics has also advanced rapidly, with machine learning-based
locomotion controllers enabling increasingly natural and complex walking behaviors [5]-[11]. In terms of mechanical
design, however, most bipedal robots still employ simple foot structures that prioritize mechanical simplicity and
durability. For instance, Cassie has a single-DOF ankle [10], and Digit has a two-DOF ankle [12]. However, neither
incorporates an active toe joint, limiting foot-ground adaptability.

Although research on robotic toes has been ongoing for decades, with studies demonstrating benefits in stride length,
shock absorption, and gait naturalness [13]-[16], the practical implementation of active toe mechanisms remains limited
due to increased complexity and structural fragility [17]. As alternatives, simpler approaches such as passive and
spring-loaded toe mechanisms have been explored, improving locomotion efficiency and terrain adaptability to
some extent [18][19]. However, these approaches lack the dynamic force modulation seen in human toes.

\begin{figure}[t]
\centering
\begin{subfigure}[b]{0.45\columnwidth}
    \centering
    \includegraphics[width=\linewidth]{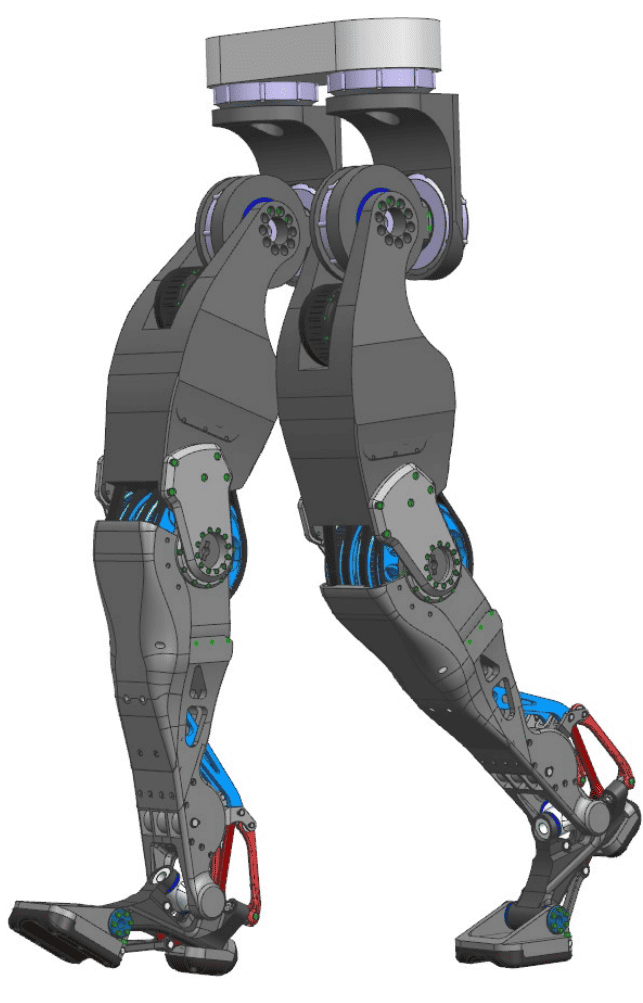}
    \subcaption{}\label{fig:robot_cad}
\end{subfigure}\hspace{0.04\columnwidth}%
\begin{minipage}[b]{0.45\columnwidth}
    \centering
    \begin{subfigure}{\linewidth}
        \centering
        \includegraphics[width=\linewidth]{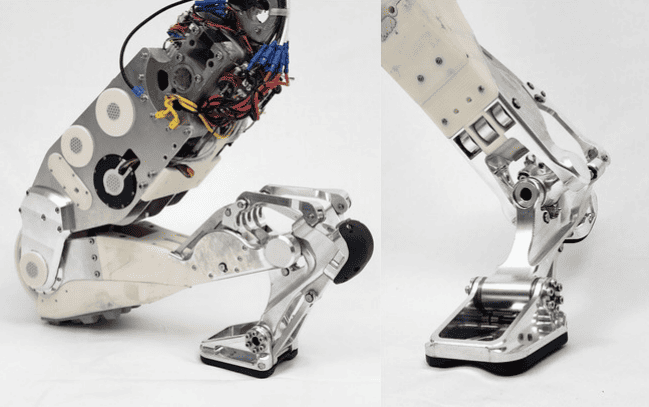}
        \subcaption{}\label{fig:hyperleg}
    \end{subfigure}\\
    \begin{subfigure}{\linewidth}
        \centering
        \includegraphics[width=\linewidth]{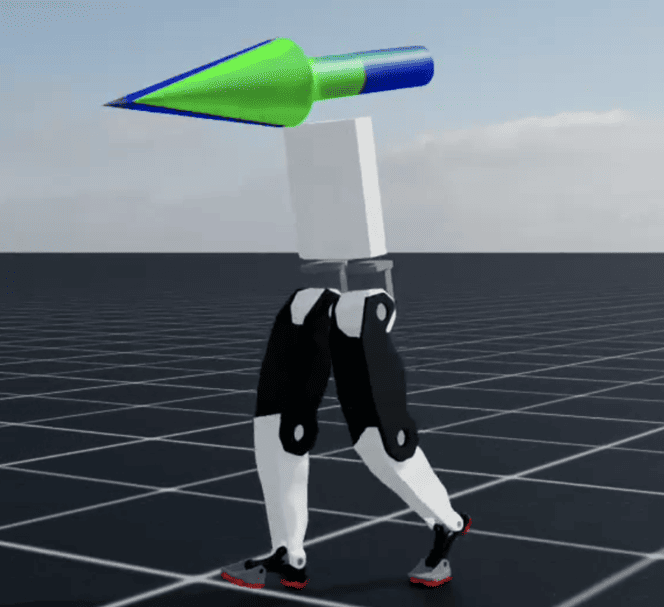}
        \subcaption{}\label{fig:simmodel}
    \end{subfigure}
\end{minipage}
\caption{(a) 14-DOF biped robot design. 
(b) HyperLeg mechanism: 2-DOF ankle, 1-DOF toe. 
(c) Simulation model matching the actual robot.}
\label{fig:robot_overview}
\end{figure}

To address these limitations, we previously developed HyperLeg, a leg mechanism featuring two DOF in the ankle and one
DOF in the toe, as shown in Fig. 1(b) [20]. This design introduced a novel toe actuation system that mimics the high
torque and lightweight structure of the human foot, providing enhanced impact resilience and agility. Building upon this foundation, this study presents a 14-DOF bipedal robot with active toes, as depicted in Fig. 1(a). The proposed leg design ensures a realistic weight distribution suitable for human-scale bipedal locomotion. Specifically, by housing all heavy actuators within the thigh frames, the mass of the distal part is minimized, maintaining a lightweight and agile leg structure despite the addition of the active toe mechanism. However, matching human size and mass alone does not guarantee human-level efficiency. An adult weighing approximately 70 kg, walking at 1.33 m/s, consumes 125 W [1], corresponding to a CoT of 0.316. Traditional bipedal robots, in contrast, exhibit CoT values typically well above 1 [21][22]. Quantifying how much the active toe mechanism can contribute to closing this efficiency gap is a central goal of this study.

The remainder of this paper is organized as follows: Section II presents the detailed design of the proposed bipedal robot and provides details on the developed actuator. Section III describes the simulation formulation and Section IV reports the simulation results with an in-depth analysis. Finally, Section V concludes the paper.

\begin{figure}[t]
    \centering
    \begin{subfigure}[b]{0.41\columnwidth}
        \centering
        \includegraphics[width=\linewidth]{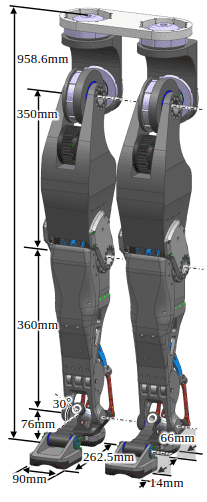}
        \subcaption{}\label{fig:biped_dim}
    \end{subfigure}\hfill
    \begin{subfigure}[b]{0.41\columnwidth}
        \centering
        \includegraphics[width=\linewidth]{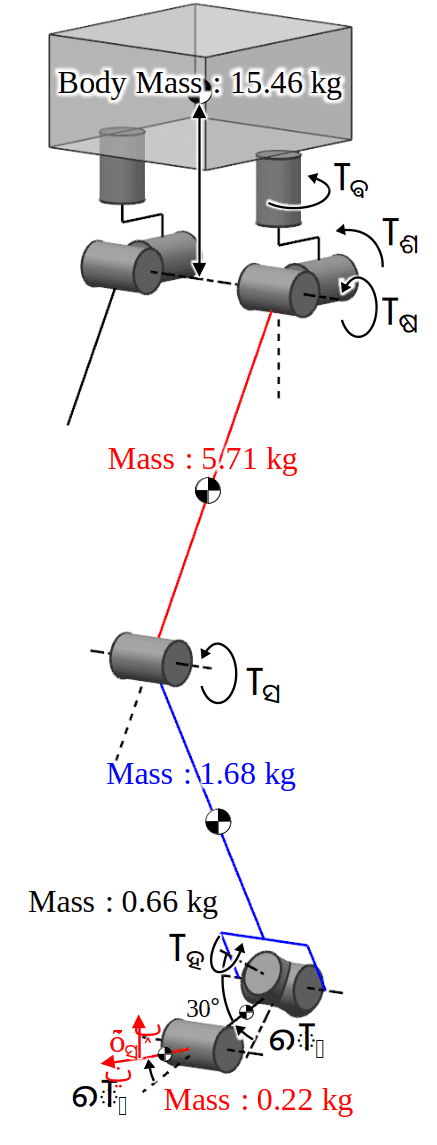}
        \subcaption{}\label{fig:coords_mass}
    \end{subfigure}

    \begin{subfigure}[b]{0.95\columnwidth}
        \centering
        \includegraphics[width=\linewidth]{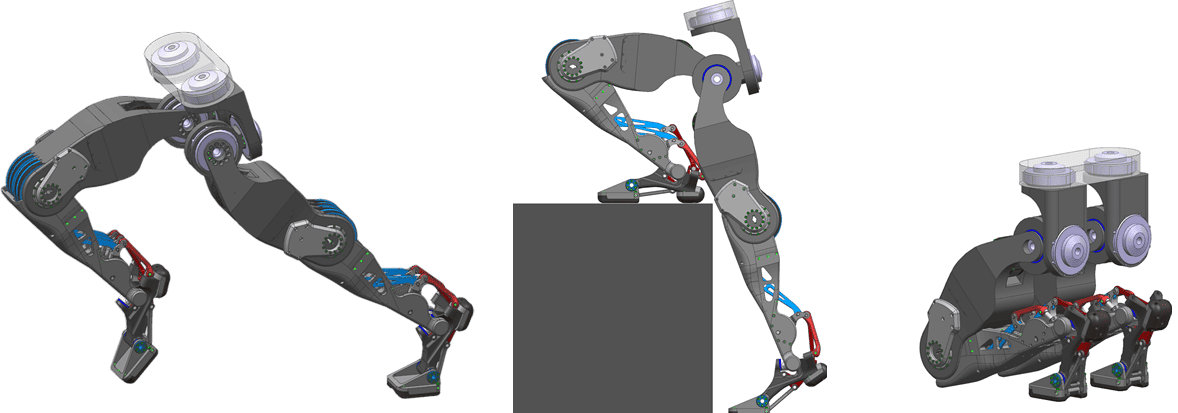}
        \subcaption{}\label{fig:motion_range}
    \end{subfigure}

    \begin{subfigure}[b]{0.31\columnwidth}
        \centering
        \includegraphics[width=\linewidth]{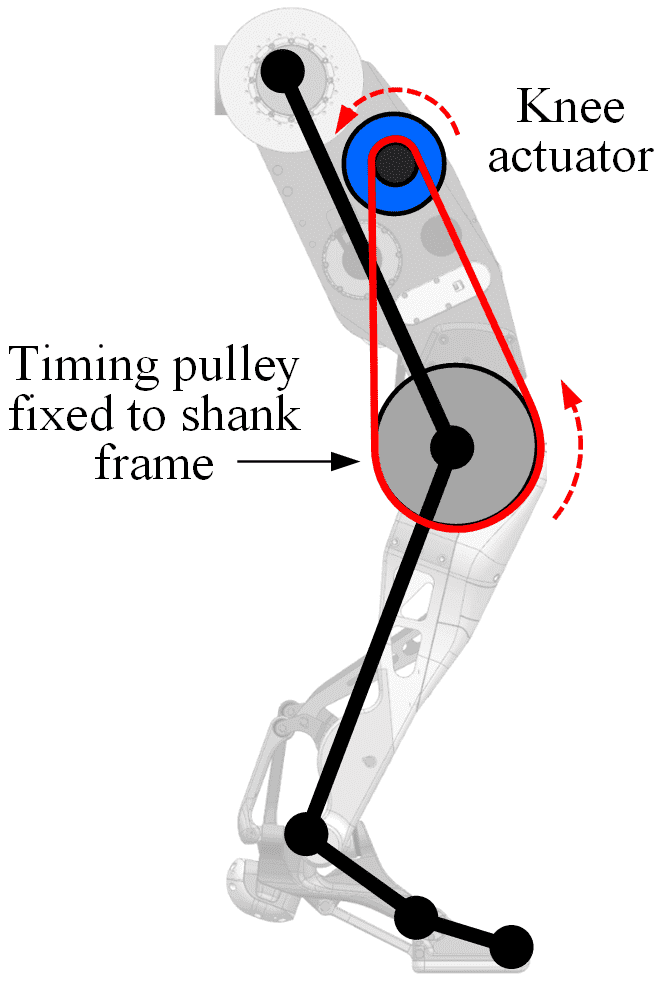}
        \subcaption{}\label{fig:knee_act}
    \end{subfigure}\hfill
    \begin{subfigure}[b]{0.31\columnwidth}
        \centering
        \includegraphics[width=\linewidth]{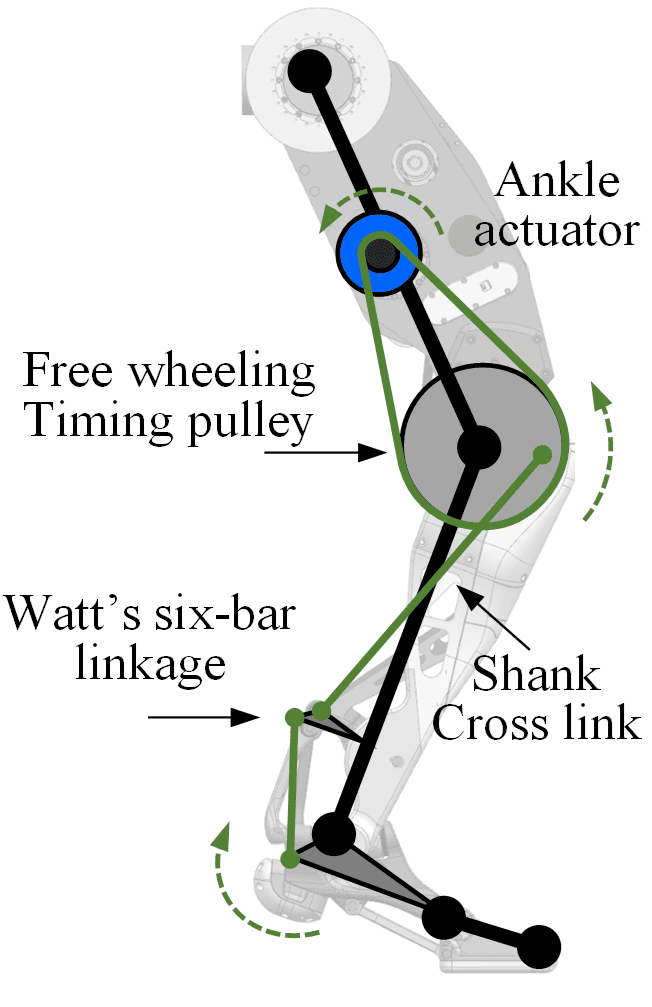}
        \subcaption{}\label{fig:ankle_act}
    \end{subfigure}\hfill
    \begin{subfigure}[b]{0.31\columnwidth}
        \centering
        \includegraphics[width=\linewidth]{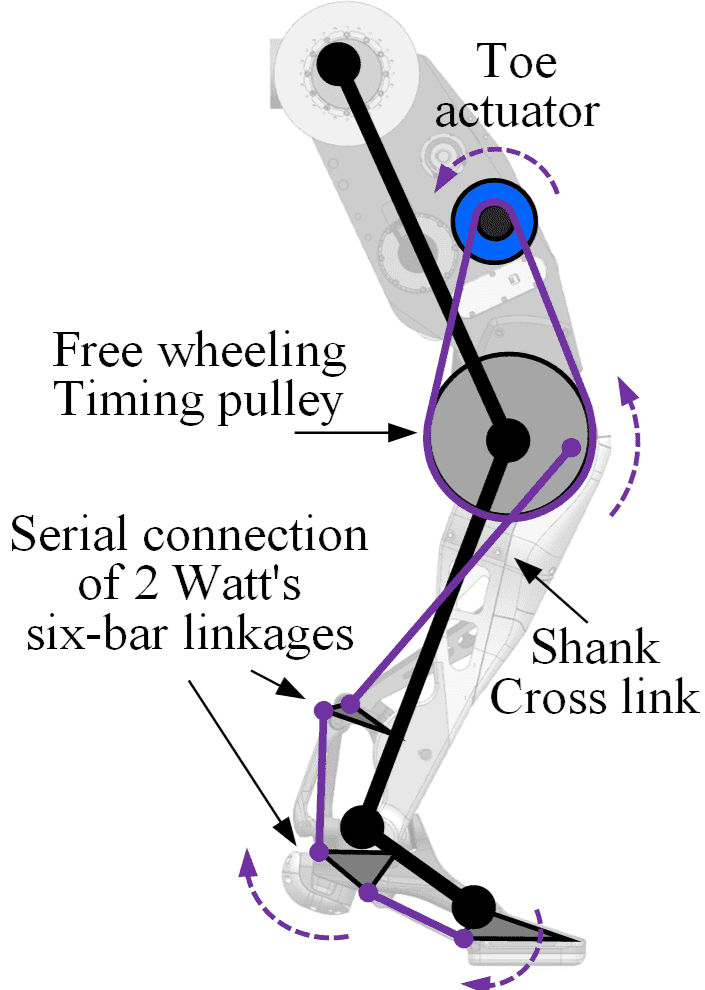}
        \subcaption{}\label{fig:toe_act}
    \end{subfigure}
    
    \caption{(a) Dimensions of the 14-DOF biped robot. 
    (b)  Coordinate frames and link masses. 
    (c) Range of motion. 
    (d)-(f) Coupled belt-linkage actuation for the knee, ankle, and toe.}
    \label{fig:hyperleg_design}
\end{figure}

\begin{figure}[t]
    \centering
    \begin{subfigure}[b]{0.8\columnwidth}
        \centering
        \includegraphics[width=\linewidth]{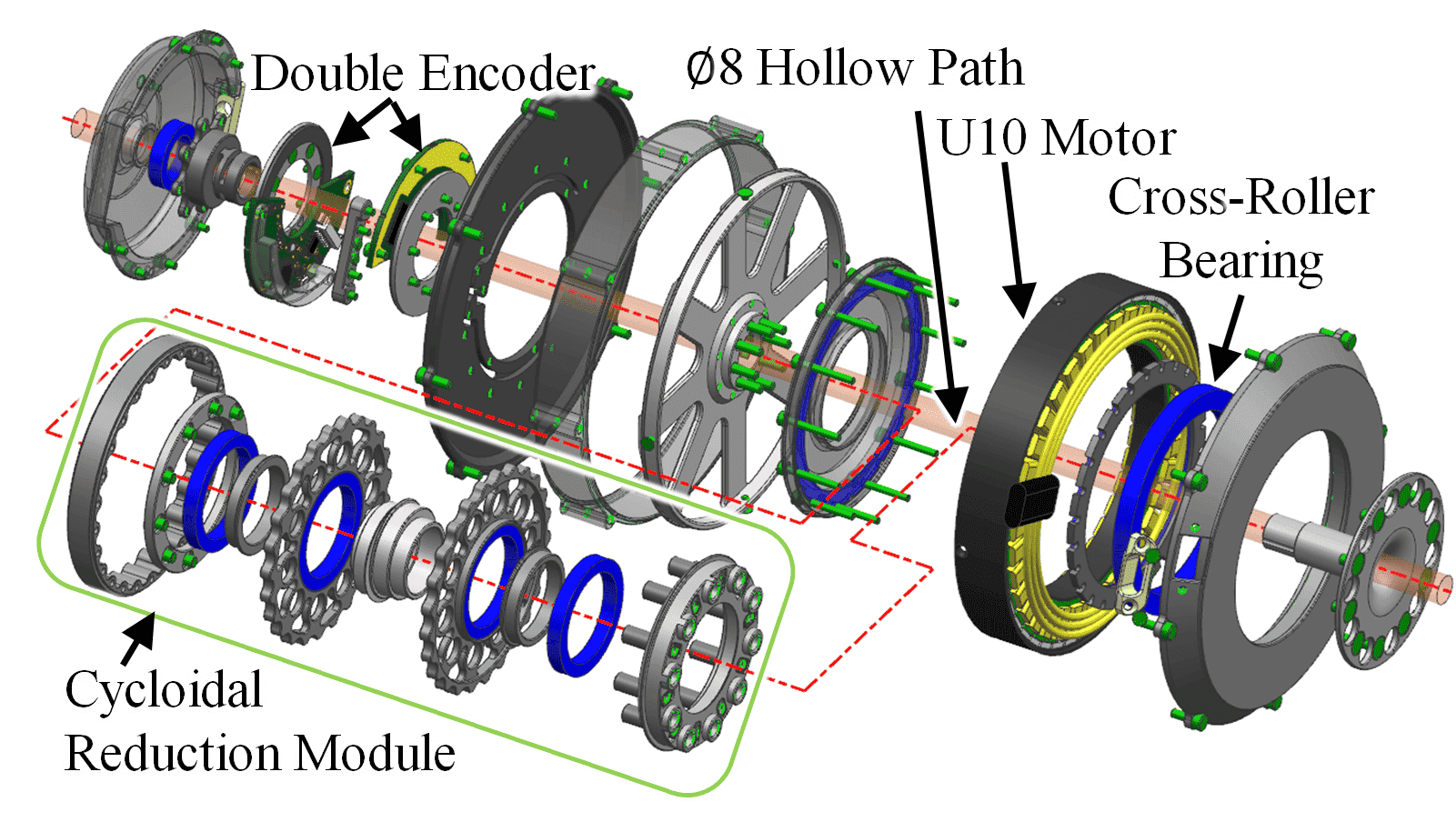}
        \subcaption{}\label{fig:hip_act}
    \end{subfigure}

    \begin{subfigure}[b]{0.5\columnwidth}
        \centering
        \includegraphics[width=\linewidth]{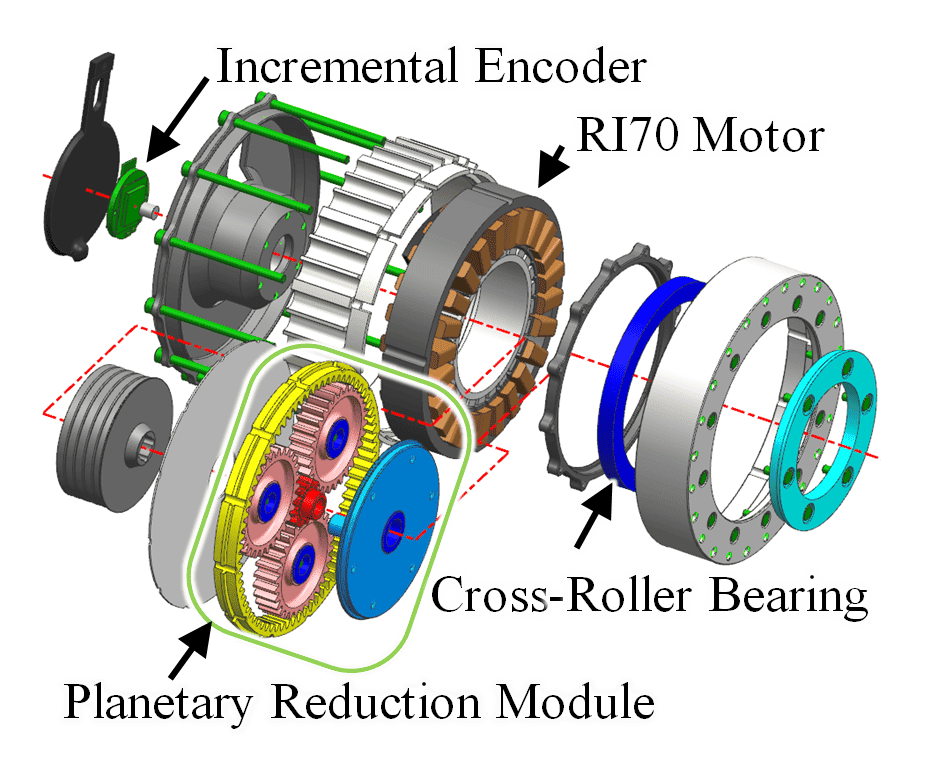}
        \subcaption{}\label{fig:knee_act_detail}
    \end{subfigure}\hfill
    \begin{subfigure}[b]{0.5\columnwidth}
        \centering
        \includegraphics[width=\linewidth]{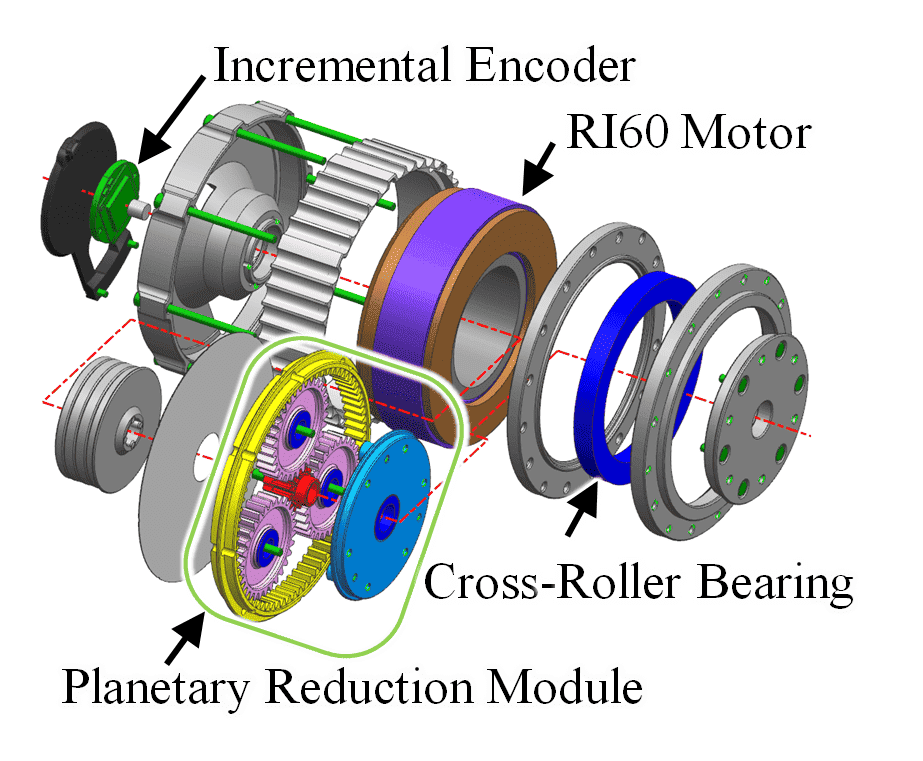}
        \subcaption{}\label{fig:ankletoe_act_detail}
    \end{subfigure}

    \begin{subfigure}[b]{0.9\columnwidth}
        \centering
        \includegraphics[width=\linewidth]{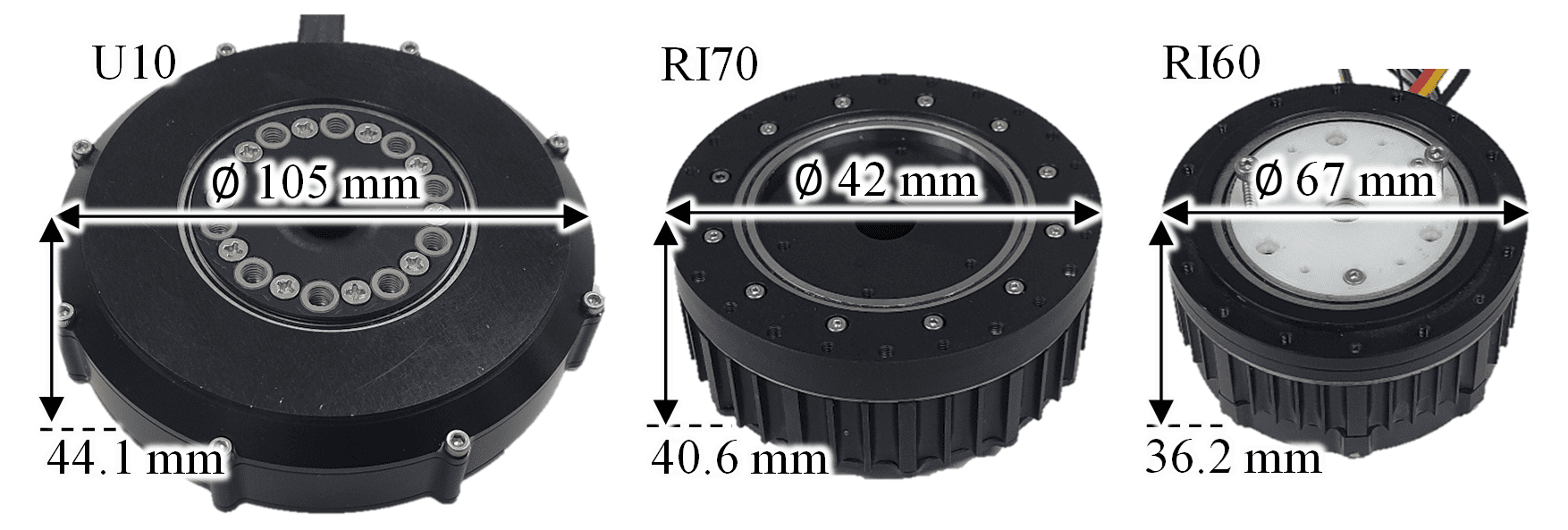}
        \subcaption{}\label{fig:impl_act}
    \end{subfigure}

    \caption{(a) Hip actuator: U10 + cycloidal gear. 
    (b)/(c) Knee and ankle/toe actuators: BLDC + planetary gear. 
    (d) Implemented actuators.}
    \label{fig:actuators}
\end{figure}
    
\section{Hardware}
\subsection{Mechanical Design}
As shown in Fig. 2(a) and (b), the bipedal robot consists of a hip with three degrees of freedom (DOF), 
a knee with one DOF, an ankle with two DOF, and a toe with one DOF. The leg length up to the hip pitch 
joint is 786 mm, and the total height is 958.6 mm. Each leg weighs 8.27 kg, and since all heavy actuators
are housed within the thigh frames, the leg features lightweight distal parts despite the active toe,
similar to those of humans, making it well-suited for agile and natural walking (Fig. 2(b)); moreover,
power transmission via belts and linkages keeps the joints compact, enabling a wide range of motion
(Fig. 2(c)). Figs. 2(d), (e), and (f) illustrate the actuators and power transmission mechanisms driving
the knee, ankle, and toe, respectively. Since the ankle has two DOF, the robot uses two identical structures 
shown in Fig. 2(e) in parallel, enabling differential actuation for ankle roll and pitch. See [20] for further details.

Fig. 3 illustrates the structure and actual images of the three
types of actuators developed. As shown in Fig. 3(a), the
actuator for the hip requires high torque output and is
therefore implemented using a 25:1 cycloid gear and an
outrunner frameless motor with a large air gap diameter. The
structures of the actuators for the knee and those for the ankle
and toe are shown in Figs. 3(b) and (c), respectively. Both are
composed of a 7:1 planetary gear and an inrunner BLDC
motor with low rotor inertia, achieving additional reduction
ratios through the power transmission mechanism. All
components were carefully selected and designed with
considerations for low friction resistance, high
backdrivability, and high-impact resistance while maintaining
high power output. These features play a crucial role in
minimizing the sim-to-real gap. The detailed specifications of
the robot and actuators are summarized in Table I.

\subsection{Cooperative Actuation}
As shown in Figs. 2(d)-(f), the knee, ankle, and toe actuators are intentionally coupled:
when the foot contacts the ground, they cooperate to extend the knee and perform ankle plantar flexion,
generating substantial propulsive force.

The Jacobian $J_m$, which represents the velocity relationship between the motors and joints, is expressed as follows:
\begin{equation}
    \dot{\theta} = J_m \, \dot{\phi},
    \label{eq:jacobian}
\end{equation}
where $\phi, \theta \in \mathbf{R}^{7}$ are the motor and joint angle vectors, respectively.
$\phi$ contains the hip yaw, roll, and pitch motors, followed by the knee, left and right ankle, 
and toe motors. $\theta$ contains the hip yaw, roll, and pitch joints, followed by the knee, ankle pitch and roll, 
and toe joints.

The relationship between the motor torques $\tau_m$ and joint torques $\tau_{jt}$ can be expressed as:
\begin{equation}
    \tau_m = J_m^T \, \tau_{jt}.
    \label{eq:torque}
\end{equation}
An example of the motor torque and joint torque relationship in a walking posture is given by the following equation:
\begin{equation}
    \resizebox{0.92\columnwidth}{!}{$
    \setlength{\arraycolsep}{2pt}
    \tau_m =
    \begin{bmatrix}
    25 & 0 & 0 & 0 & 0 & 0 & 0 \\
    0 & 25 & 0 & 0 & 0 & 0 & 0 \\
    0 & 0 & 25 & 0 & 0 & 0 & 0 \\
    0 & 0 & 0 & -31.11 & -31.11 & -31.11 & -31.11 \\
    0 & 0 & 0 & 0 & 26.69 & 26.69 & 19.23 \\
    0 & 0 & 0 & 0 & 16.51 & -16.51 & 0 \\
    0 & 0 & 0 & 0 & 0 & 0 & 21.27 \\
    \end{bmatrix}
    \tau_{jt}
    $}.
    \label{eq:walking_example}
\end{equation}
    
The hip's serial connection yields a diagonal 3$\times$3 upper-left block. 
The knee row (4th) couples all knee, ankle, and toe actuators at 31.11:1 
reduction each, and the fifth row similarly shows ankle/toe cooperation contributing 
to ankle pitch. This intentional coupling is termed Cooperative Actuation (CA) [23].

\section{Learning-Based Control}
\subsection{Simulation Setup Minimizing Sim-To-Real Gap}
Although learning-based control has recently overcome the limitations of traditional model-based control 
in locomotion research, inaccurate robot dynamics and actuator models in simulators cause significant 
sim-to-real gaps. In particular, common simulators model robots as direct-joint-actuation systems, failing to 
faithfully reproduce parallel-linkage ankle structures [12] and CA structures like HyperLeg, thereby degrading 
real-to-sim fidelity---the accuracy with which real hardware is reflected in simulation. Furthermore, reward 
functions designed for energy-efficient locomotion further simplify energy consumption by 
summing squared torques instead of using an accurate CoT. Bypassing this bottleneck by designing simpler hardware 
reduces the sim-to-real gap but forfeits hardware advantages, limiting potential performance gains.

\begin{figure}[t]
    \centering
    \begin{subfigure}[b]{0.48\columnwidth}
        \centering
        \includegraphics[width=\linewidth]{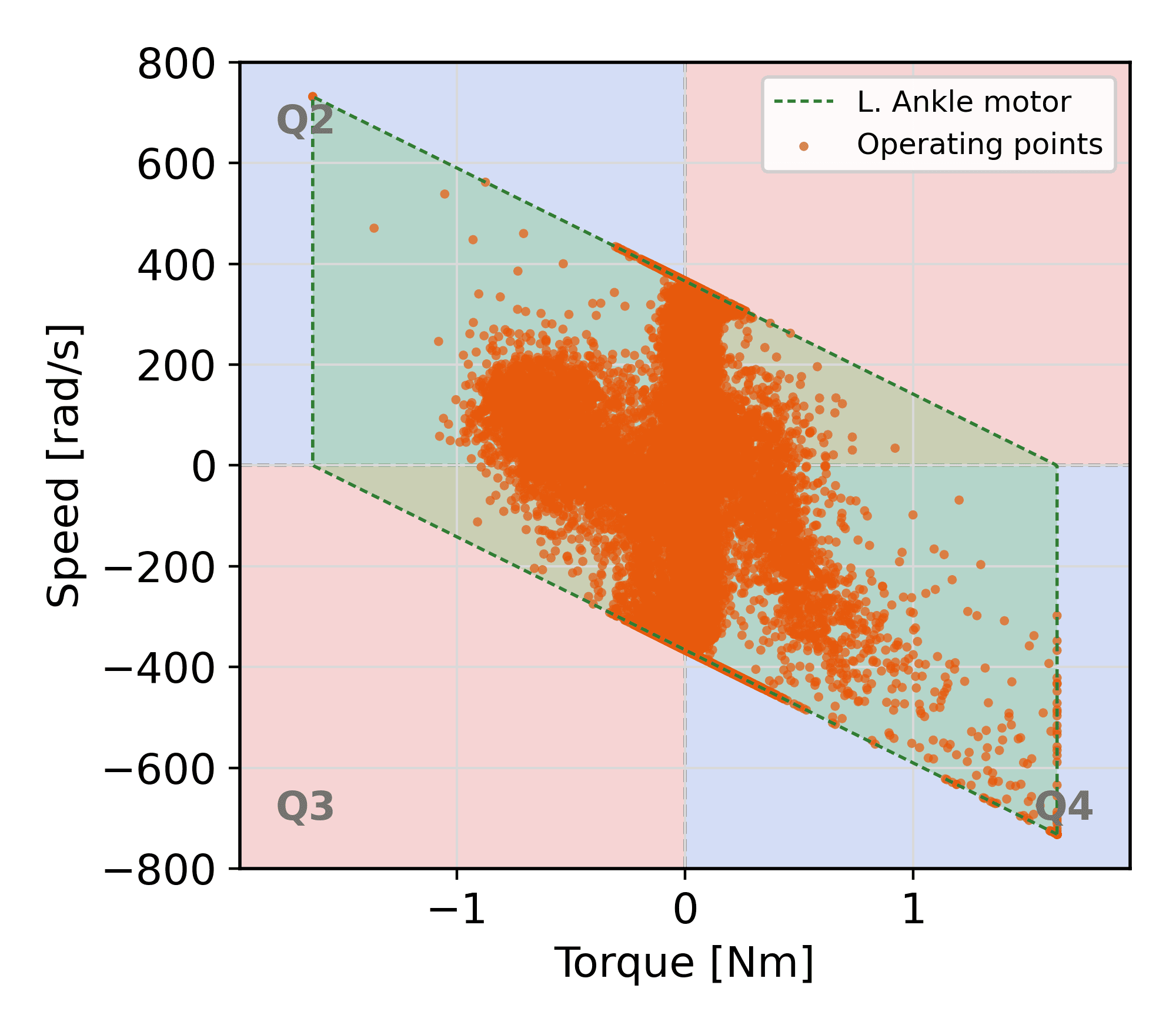}
        \subcaption{}\label{fig:env_motor_L}
    \end{subfigure}\hfill
    \begin{subfigure}[b]{0.48\columnwidth}
        \centering
        \includegraphics[width=\linewidth]{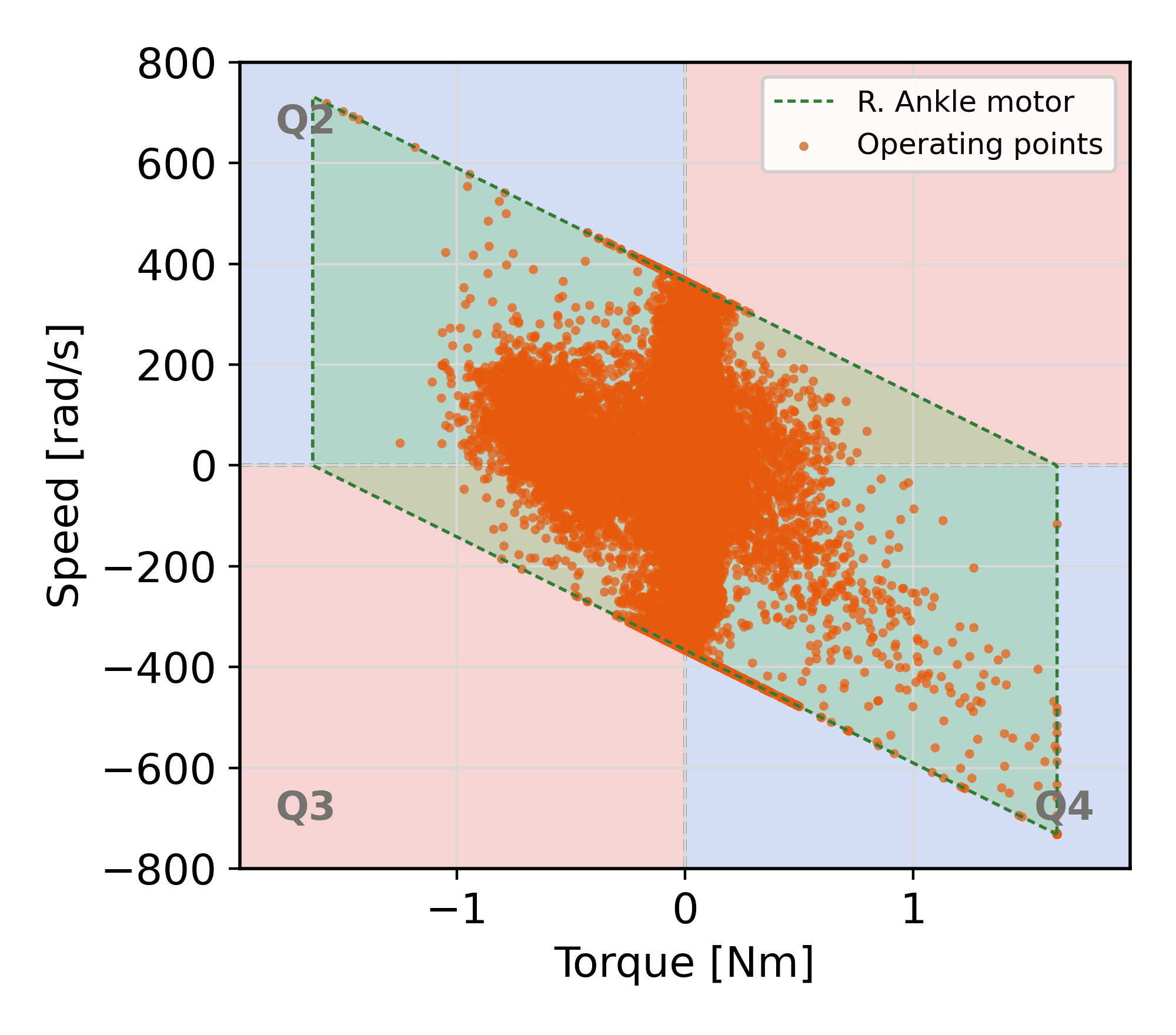}
        \subcaption{}\label{fig:env_motor_R}
    \end{subfigure}

    \begin{subfigure}[b]{0.48\columnwidth}
        \centering
        \includegraphics[width=\linewidth]{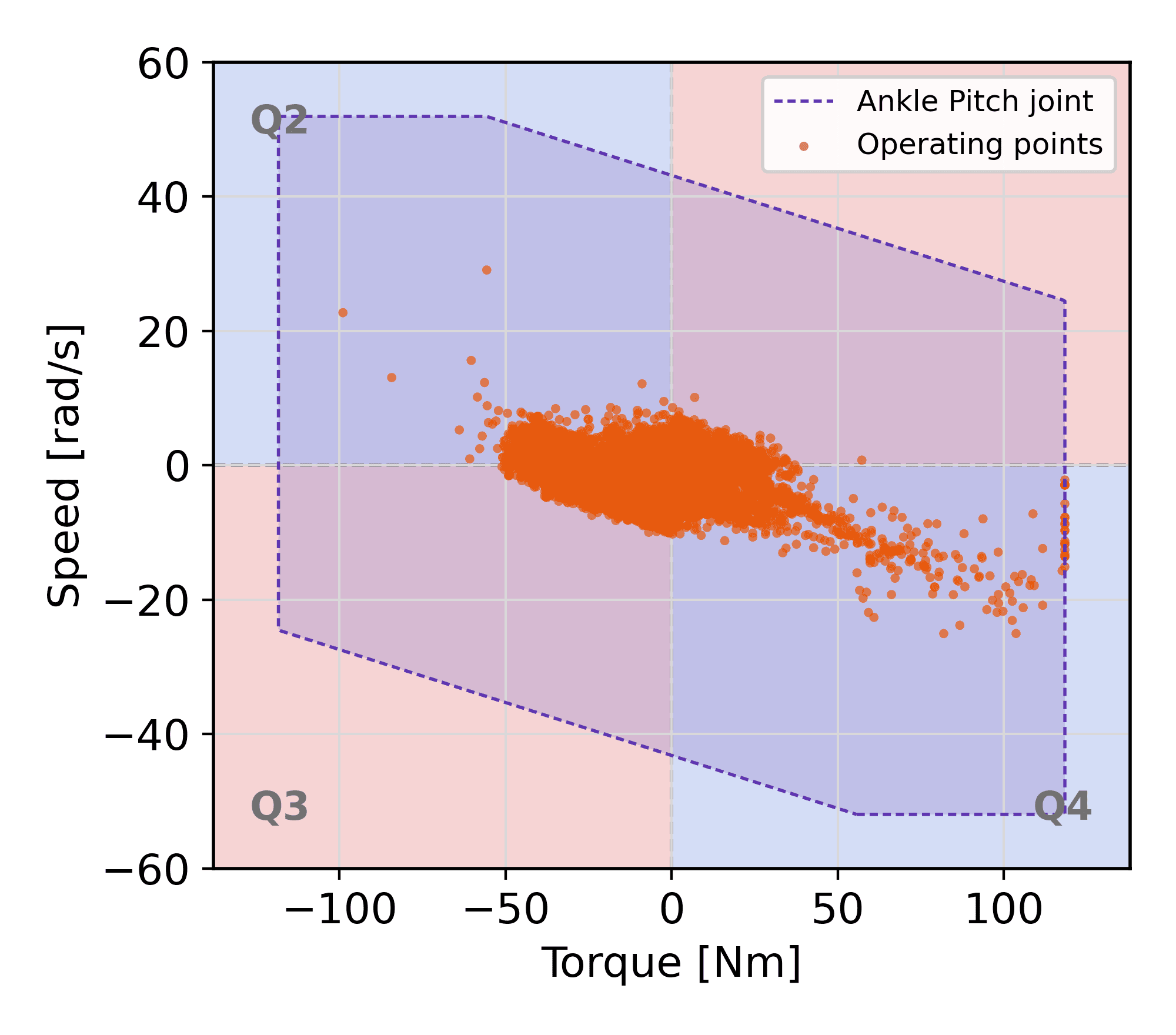}
        \subcaption{}\label{fig:env_joint_pitch}
    \end{subfigure}\hfill
    \begin{subfigure}[b]{0.48\columnwidth}
        \centering
        \includegraphics[width=\linewidth]{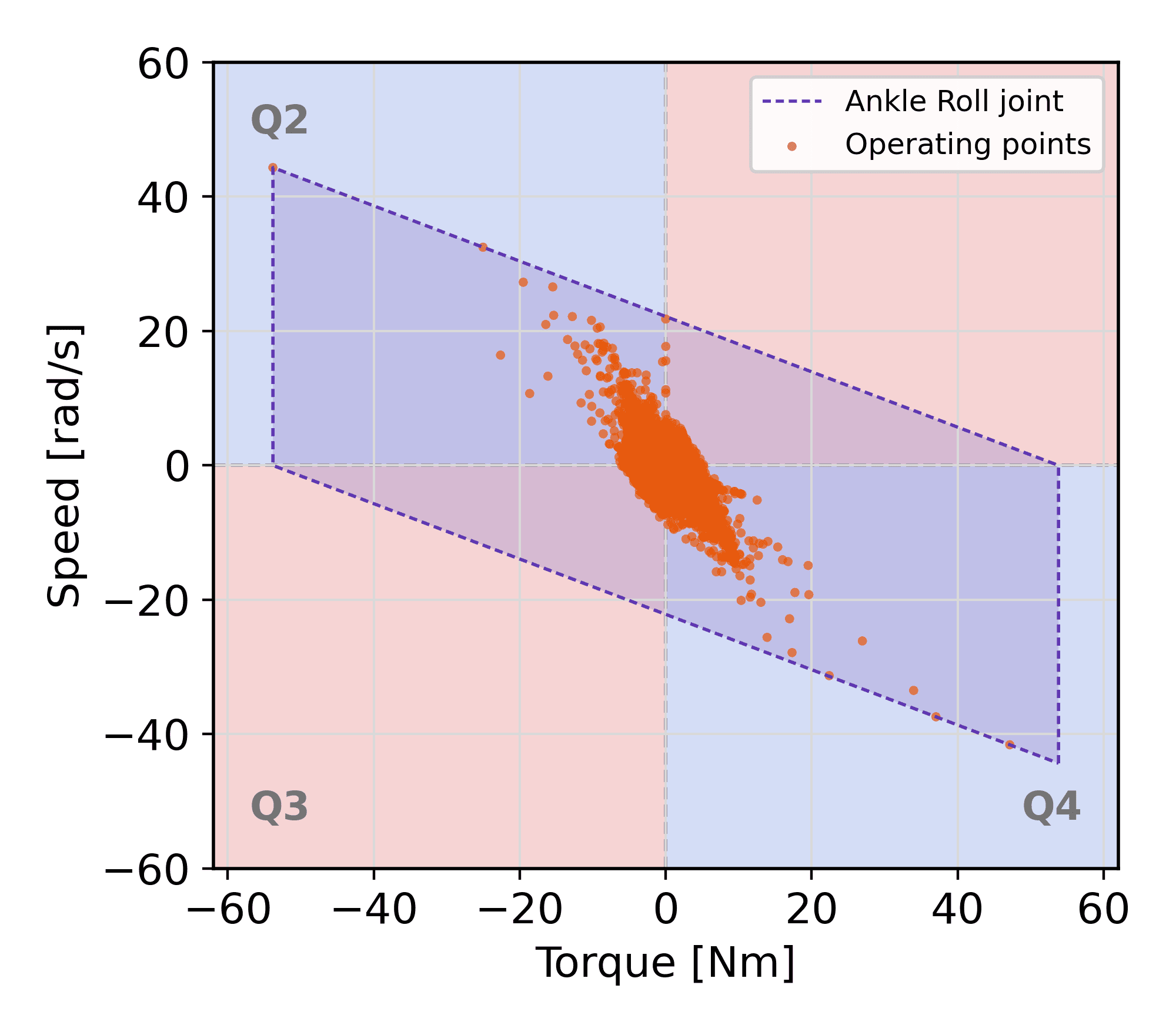}
        \subcaption{}\label{fig:env_joint_roll}
    \end{subfigure}

    \caption{Torque--speed envelopes of the CA mapping.
    (a)/(b) Left/right ankle motors.
    (c)/(d) Ankle pitch ($\mathrm{L}{+}\mathrm{R}$) and roll ($\mathrm{L}{-}\mathrm{R}$) joints, projected through the CA Jacobian.
    Red: motoring ($\tau\omega \geq 0$); Blue: regenerative ($\tau\omega < 0$).}
    \label{fig:ca_envelopes}
\end{figure}

To address these challenges, this paper focused on:
\begin{itemize}
    \item Realistic actuator modeling that accurately reflects the relationship between joints and actuators.
    \item Integration of accurate CoT estimation into training.
\end{itemize}
By incorporating these improvements, we aim to reduce the sim-to-real gap while retaining the benefits of
complex hardware architectures, ultimately enabling robust and energy-efficient real-world bipedal locomotion.

\begin{table}[t]
\centering
\caption{Mechanical Specifications}
\label{tab:mech_spec}
\renewcommand{\arraystretch}{1.15}
\setlength{\tabcolsep}{3pt}
\scriptsize
\resizebox{\columnwidth}{!}{%
\begin{tabular}{|l|c|c|c|c|c|}
\hline
\multicolumn{6}{|l|}{\textbf{Mass [kg]:} pelvis 15.46, thigh 5.71, shank 1.68, foot 0.66, toe 0.22; total 32.0} \\
\hline
\rowcolor{gray!15}
\textbf{Joints} & \textbf{Hip}$^{\dagger}$ & \textbf{Knee} & \textbf{Ank.\,P} & \textbf{Ank.\,R} & \textbf{Toe} \\
\hline
RoM [deg]                          & --- & $0{\sim}150$ & $-56{\sim}60$ & $-30{\sim}30$ & $-65{\sim}10$ \\
\thinhline
\rowcolor{gray!7}
Reduction ratio                    & 25.00:1 & 31.11:1 & 26.00:1$^{*}$ & 16.51:1$^{*}$ & 31.93:1$^{*}$ \\
\thinhline
$\omega_{\max}$ [rad/s]            & 300.0 & 327.2 & 366.0 & 366.0 & 366.0 \\
\thinhline
\rowcolor{gray!7}
$\tau_{\max}^{\text{CA}}$ [Nm]     & 126.0 & 212.0 & 111.7 & 48.4 & 46.8 \\
\thinhline
Coulomb fric.\ [Nm]                & 3.30 & 5.59 & 2.09 & 2.20 & 0.88 \\
\thinhline
\rowcolor{gray!7}
Viscous fric.\ [Nm/(rad/s)]        & 0.08 & 0.10 & 0.10 & 0.10 & 0.10 \\
\hline
\rowcolor{gray!15}
\textbf{Motors} & \textbf{Hip} & \textbf{Knee} & \multicolumn{2}{c|}{\textbf{Ankle}} & \textbf{Toe} \\
\hline
Product                            & U10 & RI70 & \multicolumn{2}{c|}{RI60} & RI60 \\
\thinhline
\rowcolor{gray!7}
$\omega_{0}$ [rad/s]               & 300.0 & 327.2 & \multicolumn{2}{c|}{366.0} & 366.0 \\
\thinhline
$\tau_{\max}$ [Nm]                 & 5.04 & 2.68 & \multicolumn{2}{c|}{1.63} & 1.63 \\
\thinhline
\rowcolor{gray!7}
$C_{p}$                            & 3.54 & 33.44 & \multicolumn{2}{c|}{124.57} & 124.57 \\
\thinhline
Driver eff.\ (out / regen)         & \multicolumn{5}{c|}{0.9 / 0.8} \\
\hline
\end{tabular}%
}
\begin{flushleft}
{\scriptsize
$^{*}$Reduction at standing pose.\quad
$^{\dagger}$Hip yaw / roll / pitch share values except RoM: $-35{\sim}15$ / $-5{\sim}90$ / $-120{\sim}120$ [deg].
}
\end{flushleft}
\end{table}

Fig.~\ref{fig:ca_envelopes} shows representative operating points of the ankle motors during a simulation in
which the robot performs the agility test. Because the CA structure breaks the one-to-one mapping between motor
and joint torques, the joint velocity from the simulator and the target joint torque from the policy network are
first projected into motor space via Eqs.~(1)--(3); the motor torques are then clipped against the 4-quadrant
torque--speed envelope using the Isaac Lab DCMotorActuator [26]; finally, the clipped motor torques are
projected back to yield the CA-projected joint torque $\tau_{j\_clip}$. Since HyperLeg transmits power below
the knee through timing belts and linkages, joint-space friction is non-negligible; we therefore subtract a
measured Coulomb-plus-viscous friction term (Table~I) from $\tau_{j\_clip}$ to obtain the final commanded
torque:
\begin{equation}
    \tau_{cmd} = \tau_{j\_clip} - \left( f_{c} \cdot \tanh\!\left(\dot{q}/v_{m}\right) + f_{v}\,\dot{q} \right)
    \label{eq:friction}
\end{equation}

To rigorously isolate the contribution of the active toe to energy efficiency, the common practice of
penalizing the 2-norm of joint torques is inadequate. We instead estimate and minimize the CoT directly,
decomposing the motor-side energy consumption during locomotion into Joule heating and the (signed) mechanical
power:
\begin{equation}
    CoT = \frac{P_{Joule} + P_{mech}}{mgv}
    \label{eq:cot}
\end{equation}
where:
\begin{itemize}
    \item $m$, $g$, $v$: robot mass, gravity, and velocity.
    \item $P_{\text{Joule}}$: resistive heating $C_{p}\tau^{2}$.
    \item $P_{\text{mech}}$: signed mechanical power $\tau\omega$.
\end{itemize}

\subsection{Reinforcement Learning Setup}
To enable a fair comparison between configurations with and without the active toe, we train two variants
under an identical setup: a model with the active toe (\emph{toe-equipped}) and one with the active toe removed
(\emph{toe-ablation}). Both are given the same observation space---including all privileged simulation
signals---ensuring an unbiased setup. The policy, network, and reward-function hyperparameters are summarized
in Table~\ref{tab:rl_setup}.

However, during training we observed that the policy occasionally converged to unrealistic postures that
sustain excessive torque on specific joints. To suppress such behavior, we adopt a CBF-style soft constraint
in the same form as the recent thermal-aware reward proposed for quadruped locomotion [27]. As shown in
Table~\ref{tab:rl_setup}, $h_{i}$ denotes the safety margin of the $i$-th motor, defined as
$h_{i} = h^{\text{th}}_{i} - \mathrm{EMA}(\tau_{i}/\tau_{\text{cont},i})^{2}$, and the EMA time constant
is set to $\tau_{\text{EMA}} = 3$\,s. The per-motor thresholds $h^{\text{th}}_{i}$ are tuned heuristically
rather than derived from detailed thermal characterization, with the goal of preventing the policy from
converging to inefficient postures.

\begin{table}[t]
    \centering
    \caption{Reinforcement Learning Setup}
    \label{tab:rl_setup}
    \renewcommand{\arraystretch}{1.1}
    \setlength{\tabcolsep}{2pt}
    \scriptsize
    \begin{tabular}{|l|p{0.45\columnwidth}|>{\raggedleft\arraybackslash}p{0.1\columnwidth}|}
    \hline
    \rowcolor{gray!15}
    \multicolumn{3}{|l|}{\textbf{Algorithm \& network}} \\
    \hline
    Optimizer / framework  & \multicolumn{2}{l|}{PPO [25] / Isaac Lab [26]} \\
    Network                & \multicolumn{2}{l|}{MLP [512, 256, 128], ELU} \\
    Architecture           & \multicolumn{2}{l|}{Symmetric actor--critic} \\
    \hline
    \rowcolor{gray!15}
    \multicolumn{3}{|l|}{\textbf{Environment}} \\
    \hline
    Parallel envs / episode  & \multicolumn{2}{l|}{8192 / 20\,s} \\
    Policy / sim.\ frequency & \multicolumn{2}{l|}{50 / 200\,Hz} \\
    \hline
    \rowcolor{gray!15}
    \multicolumn{3}{|l|}{\textbf{Action \& observation}} \\
    \hline
    Action       & \multicolumn{2}{p{0.55\columnwidth}|}{EMAJointPositionAction ($\alpha = 0.2$) \newline 14 dim (w/ toe) / 12 dim (w/o toe)} \\
    Observation  & \multicolumn{2}{p{0.55\columnwidth}|}{Proprio (65 / 57 dim) \newline + Privileged (193 dim)} \\
    \hline
    \rowcolor{gray!15}
    \multicolumn{3}{|l|}{\textbf{Training (PPO)}} \\
    \hline
    Iterations / steps per rollout        & \multicolumn{2}{l|}{3000 / 24} \\
    Learning rate                         & \multicolumn{2}{l|}{$1{\times}10^{-4}$, adaptive (KL 0.01)} \\
    $\gamma$ / $\lambda$ / clip / entropy & \multicolumn{2}{l|}{0.99 / 0.95 / 0.2 / 0.01} \\
    \hline
    \rowcolor{gray!15}
    \multicolumn{3}{|l|}{\textbf{Reward (walking)}} \\
    \hline
    \rowcolor{gray!7}
    \textbf{term} & \textbf{formula} & \multicolumn{1}{c|}{\textbf{$w$}} \\
    \hline
    Termination          & $\mathds{1}[\text{terminated}]$                                                                & $-200$ \\
    Power consumption    & $P_{\text{Joule}} + P_{\text{mech}}$ (Eq.~(6))                                                 & $-10^{-4}$ \\
    Thermal penalty      & $\sum_{i} \big[-(\dot{h}_{i} + h_{i}/\tau_{\text{EMA}})\big]^{+}$                              & $-10.0$ \\
    Linear velocity (xy) & $\exp(-\|v^{\text{cmd}}_{xy} - v_{xy}\|^{2} / 0.1)$                                            & $+2.5$ \\
    Angular velocity (z) & $\exp(-(\omega^{\text{cmd}}_{z} - \omega_{z})^{2} / 0.2)$                                      & $+0.5$ \\
    \hline
    \rowcolor{gray!15}
    \multicolumn{3}{|l|}{\textbf{Reward (position-command)}} \\
    \hline
    \rowcolor{gray!7}
    \textbf{term} & \textbf{formula} & \multicolumn{1}{c|}{\textbf{$w$}} \\
    \hline
    Goal arrival      & $\mathds{1}[\|p_{xy} - p^{\text{cmd}}_{xy}\| < \sigma_{\text{arr}}]$       & $+150$ \\
    Position progress & $\max(d^{\star} - \|p_{xy} - p^{\text{cmd}}_{xy}\|, 0)$                   & $+10$ \\
    \hline
    \end{tabular}
    \end{table}

\section{Validation}

\begin{figure}[H]
    \centering
    \includegraphics[width=\linewidth]{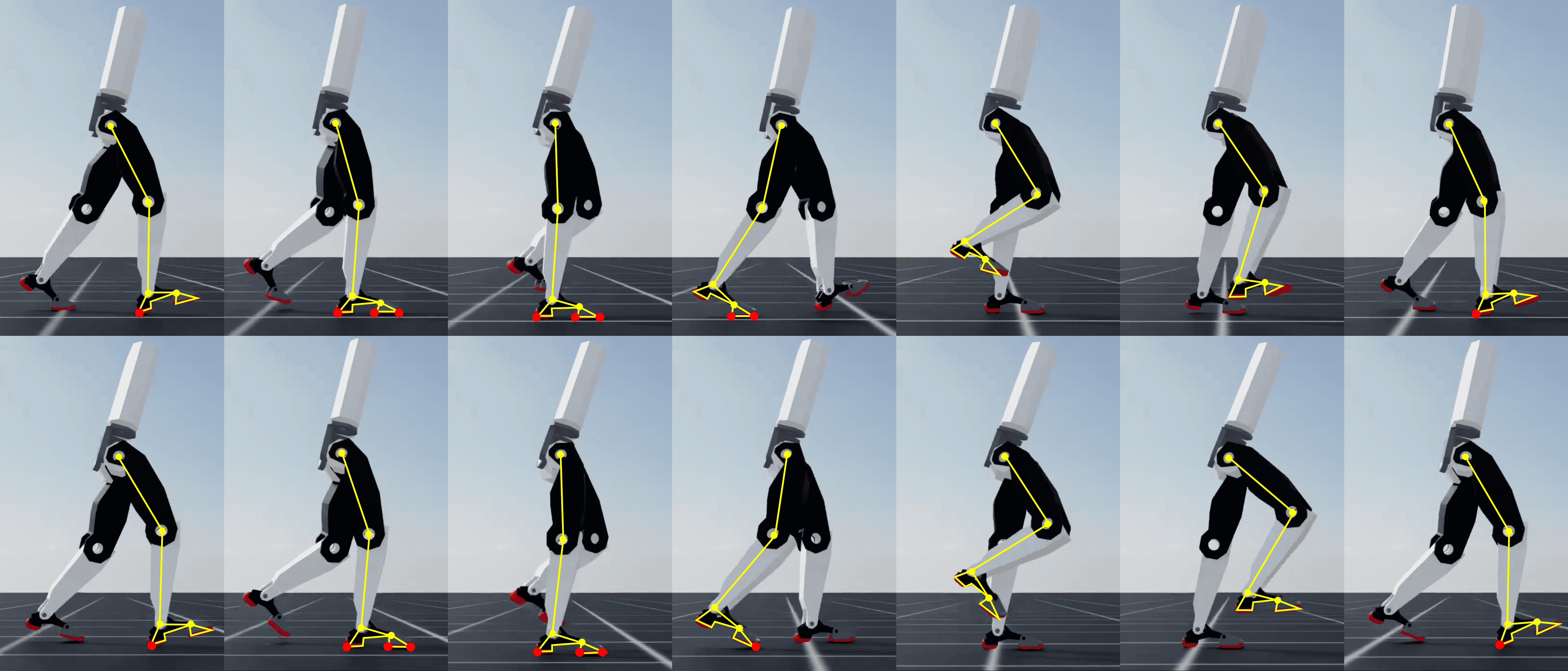}
    \caption{Gait snapshots at 1.33\,m/s. Top: toe-equipped; bottom: toe-ablation.}
    \label{fig:validation}
\end{figure}

\subsection{Efficient Walking and Impact Absorption}

To compare energy efficiency and impact absorption with and without an active toe, we performed ten
straight-line walking trials over a 20~m path at 1.33~m/s for each configuration. In Fig.~\ref{fig:validation}, the toe-ablation
model (bottom) shows greater knee flexion and leg lift than the toe-equipped model (top), implying
greater energy demand without the toe.

\begin{figure}[t]
    \centering
    \begin{subfigure}[t]{0.49\linewidth}
        \centering
        \includegraphics[width=\linewidth]{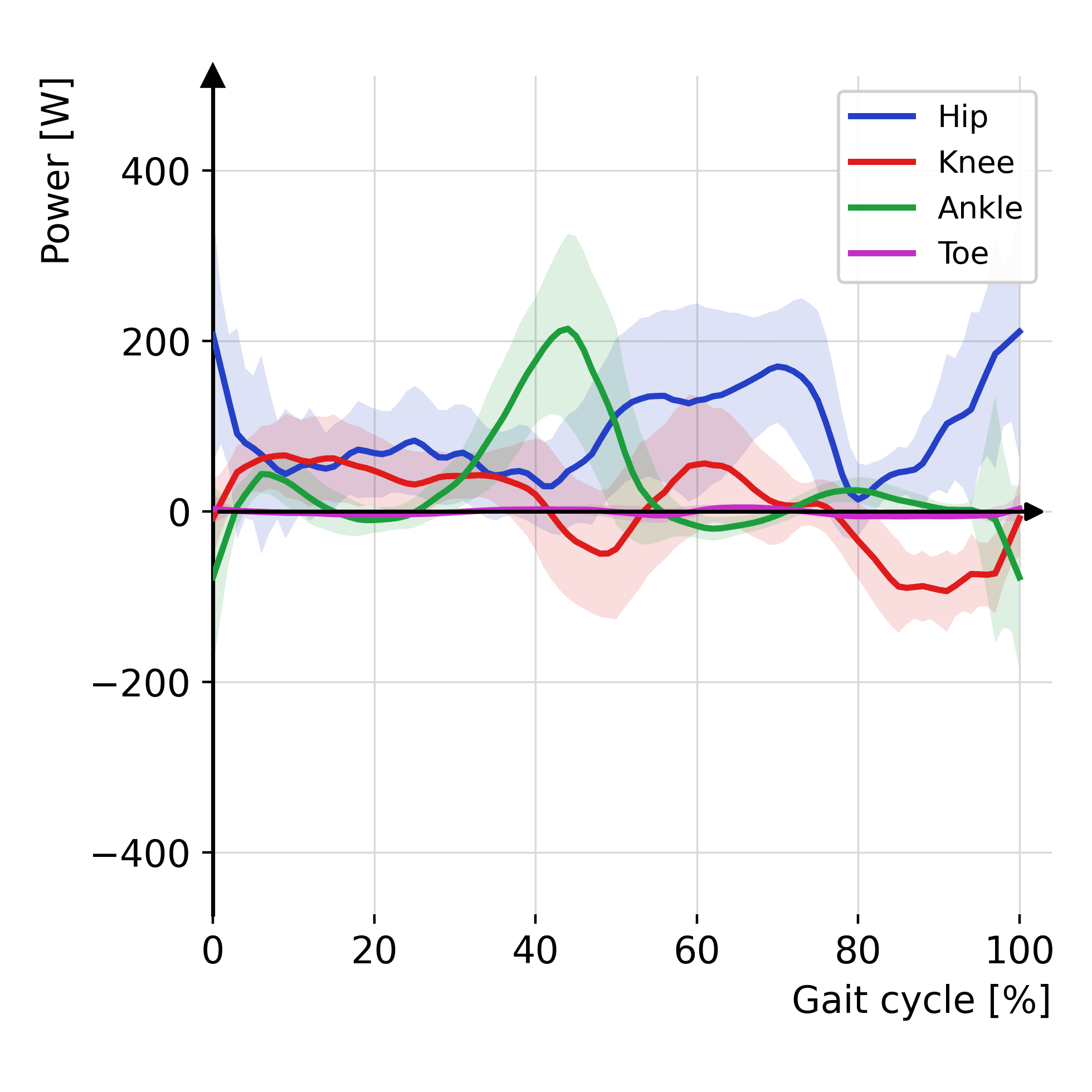}
        \subcaption{Toe-equipped}\label{fig:cyc_power_toe}
    \end{subfigure}\hfill
    \begin{subfigure}[t]{0.49\linewidth}
        \centering
        \includegraphics[width=\linewidth]{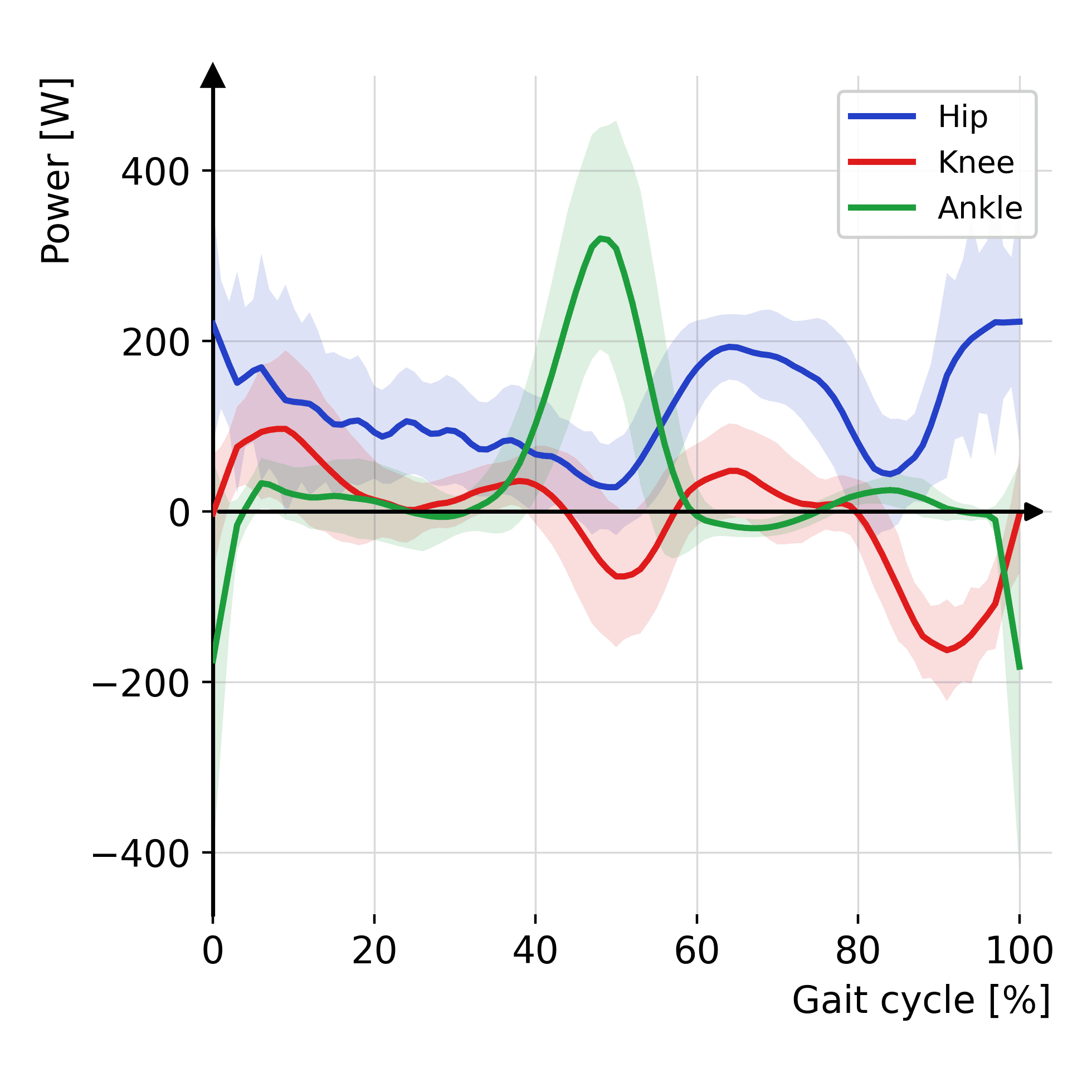}
        \subcaption{Toe-ablation}\label{fig:cyc_power_notoe}
    \end{subfigure}

    \begin{subfigure}[t]{0.49\linewidth}
        \centering
        \includegraphics[width=\linewidth]{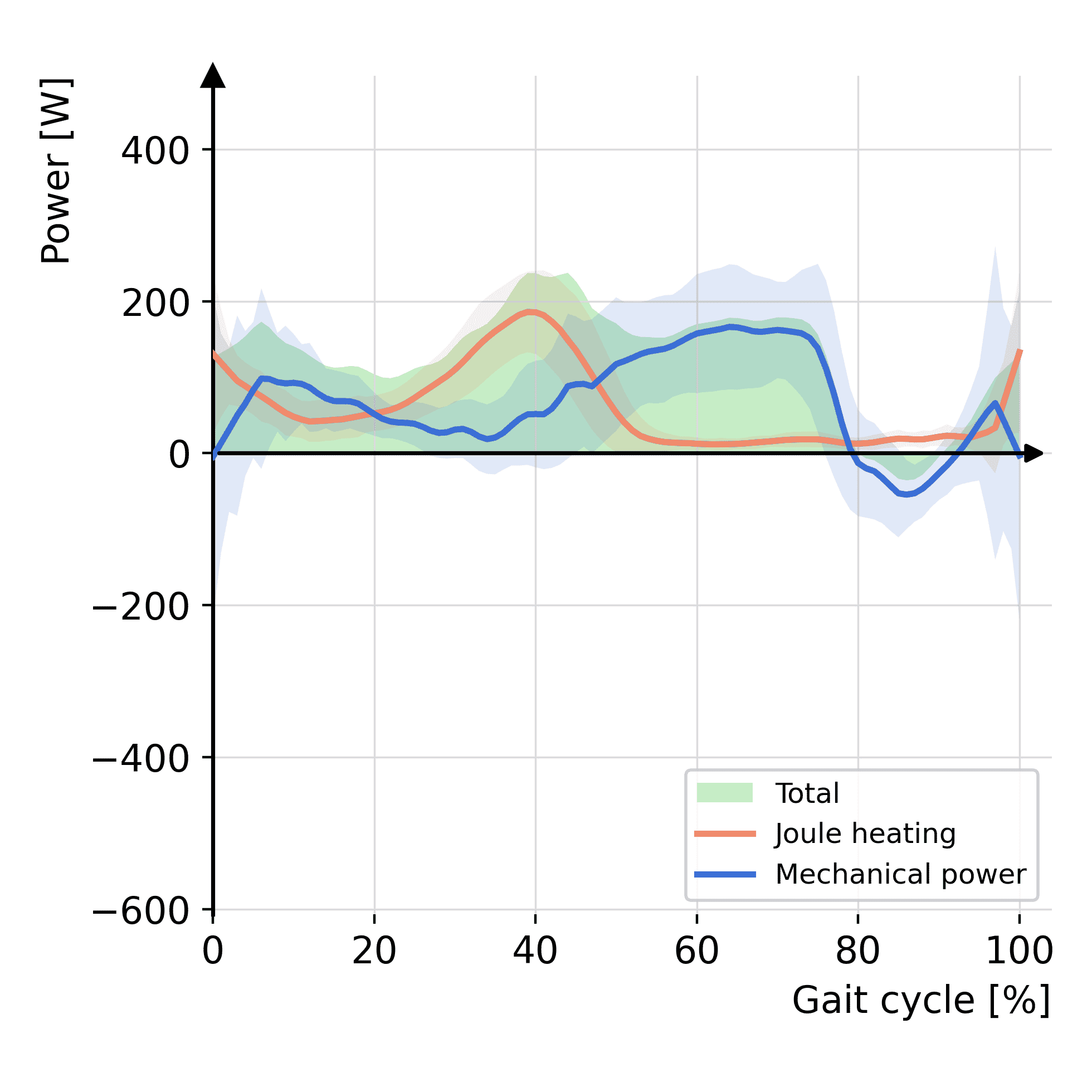}
        \subcaption{Toe-equipped}\label{fig:cyc_breakdown_toe}
    \end{subfigure}\hfill
    \begin{subfigure}[t]{0.49\linewidth}
        \centering
        \includegraphics[width=\linewidth]{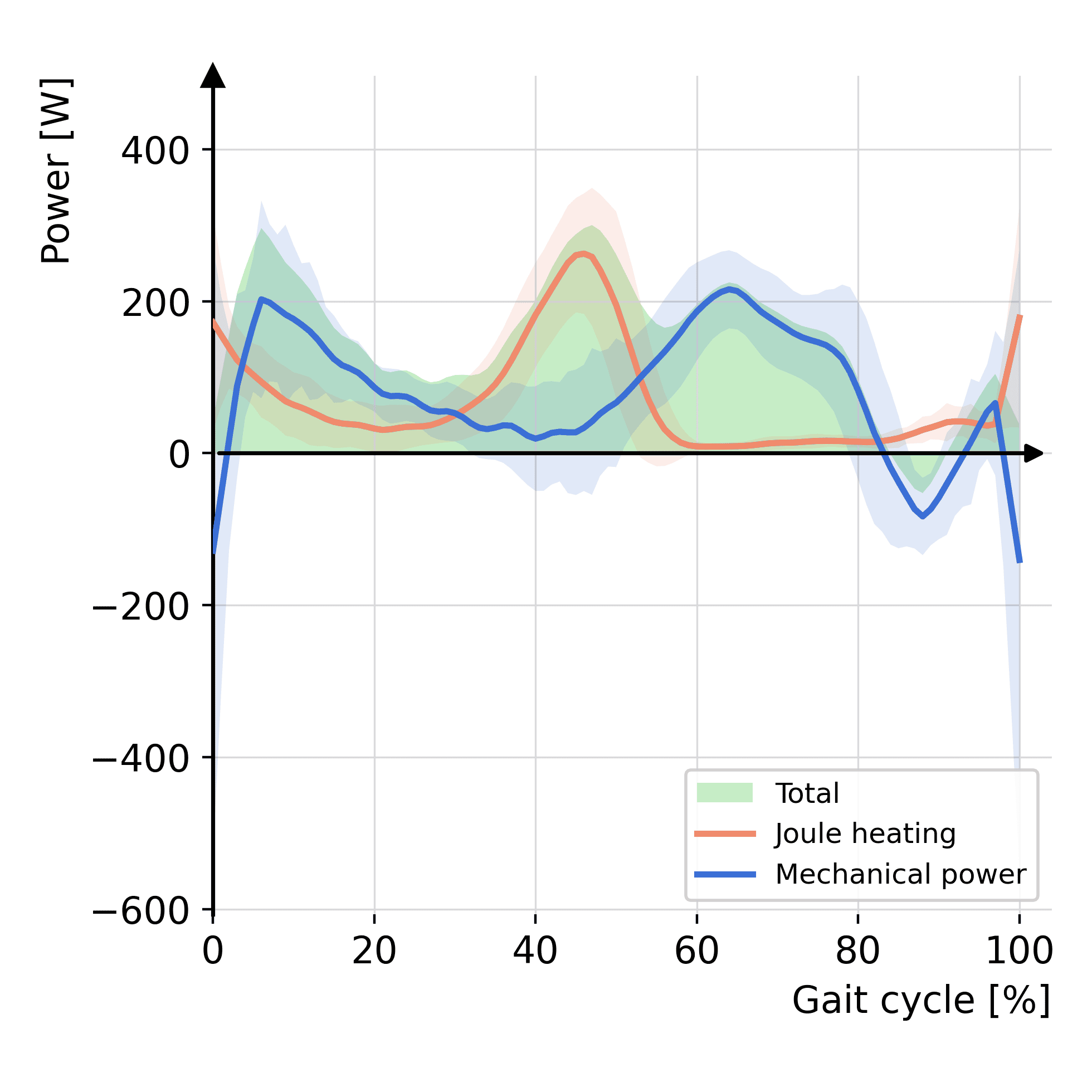}
        \subcaption{Toe-ablation}\label{fig:cyc_breakdown_notoe}
    \end{subfigure}

    \caption{Gait-cycle-averaged right-leg power per joint (top) and total breakdown (bottom).}
    \label{fig:cycle_average}
\end{figure}

Table~\ref{tab:quant_results}\footnote{$\Delta$ is the relative change of toe-equipped compared with
toe-ablation.} summarizes the whole-body metrics: relative to toe-ablation, toe-equipped reduces total power by 16.9\% (313.0~W to 260.2~W), CoT by 17.5\% (0.776 to 0.640), and heel-strike
GRF by 5.0\% (1021.1~N to 970.0~N). Fig.~\ref{fig:cycle_average}
breaks down the gait-cycle-averaged right-leg power by joint: removing the toe increases hip, knee, and ankle
power from 103.7~W to 126.0~W ($+22\%$), 61.8~W to 68.9~W ($+11\%$), and 47.5~W to 61.1~W ($+29\%$),
respectively. At the same walking speed, the toe-equipped model is superior in both energy efficiency and
impact absorption.

\begin{table}[t]
    \centering
    \caption{Walking experiment results at 1.33\,m/s}
    \label{tab:quant_results}
    \small
    \setlength{\tabcolsep}{4pt}
    \renewcommand{\arraystretch}{1.3}
    \begin{tabular}{@{}l@{\hspace{1.0em}}c@{\hspace{1.0em}}c@{\hspace{0.85em}}c@{}}
    \toprule
    \textbf{Items} & \textbf{Toe-equipped} & \textbf{Toe-ablation} & \textbf{$\Delta$ [\%]} \\
    \midrule
    \rowcolor{gray!15}
    Total Power [W]      & 260.2 & 313.0 & $-$16.9 \\
    Joule heating [W]    & 125.9 & 167.7 & $-$24.9 \\
    \rowcolor{gray!15}
    Mechanical loss [W]  & 134.3 & 145.3 & $-$7.6 \\
    CoT                  & 0.640 & 0.776 & $-$17.5 \\
    \rowcolor{gray!15}
    Time to goal [s]     & 15.96 & 16.13 & $-$1.1 \\
    Avg GRF [N]          & 970.0 & 1021.1 & $-$5.0 \\
    \bottomrule
    \end{tabular}
\end{table}

\subsection{Agility Test}

\begin{figure}[t]
    \centering
    \begin{subfigure}[t]{0.32\linewidth}
        \centering
        \includegraphics[width=\linewidth]{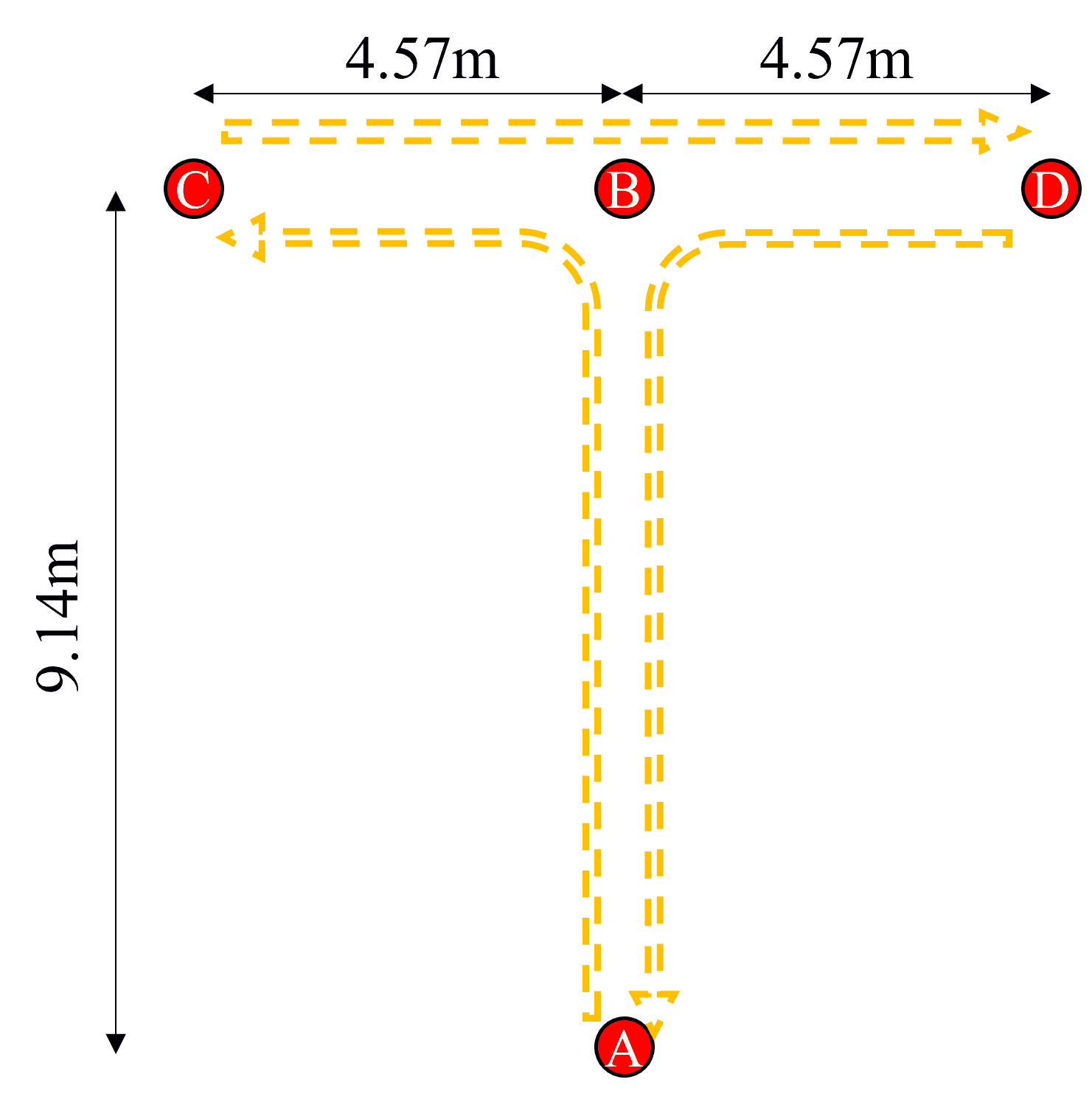}
        \subcaption{T-test layout}\label{fig:ttest_layout}
    \end{subfigure}\hfill
    \begin{subfigure}[t]{0.32\linewidth}
        \centering
        \includegraphics[width=\linewidth]{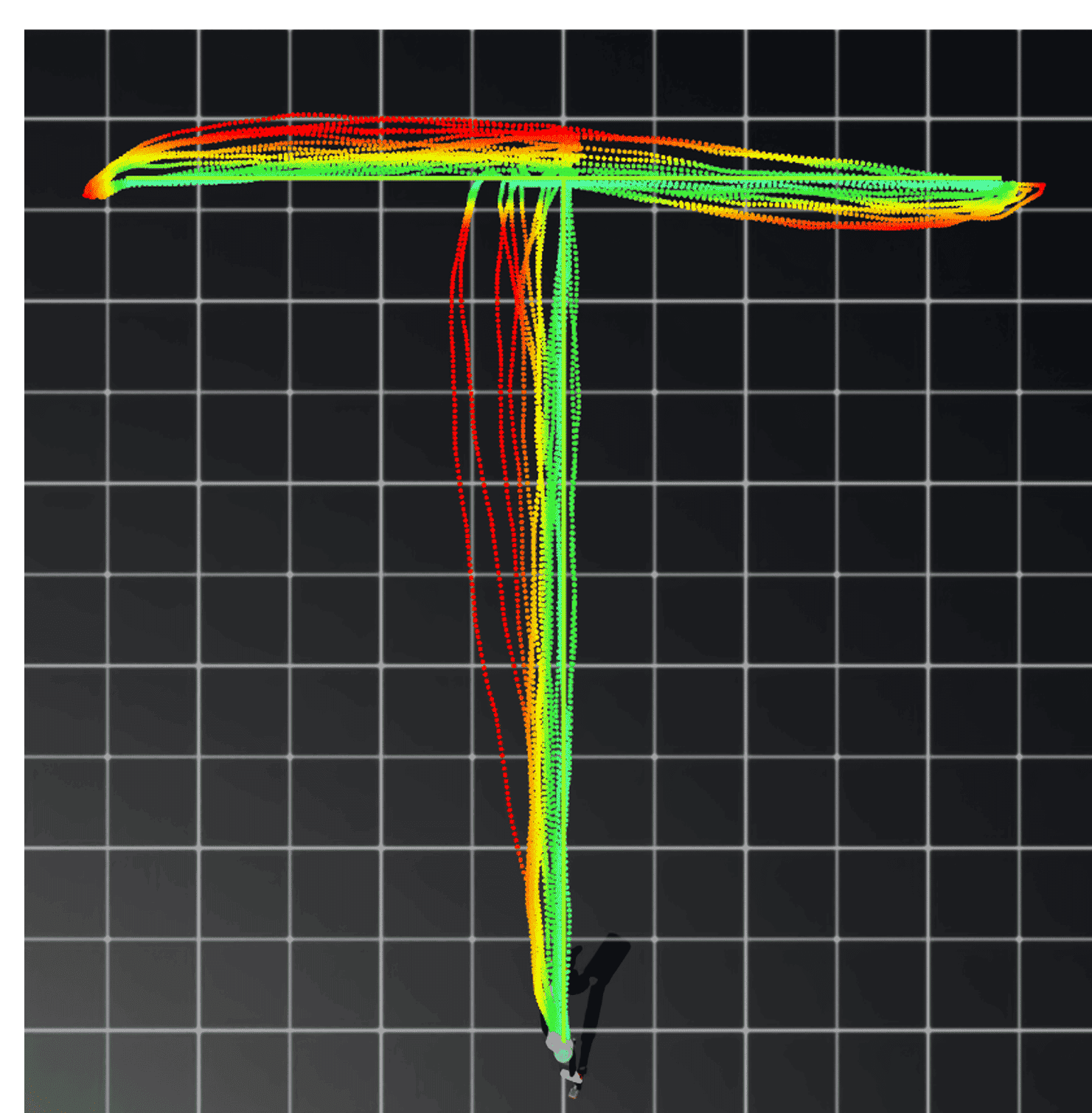}
        \subcaption{Toe-equipped}\label{fig:ttest_equipped}
    \end{subfigure}\hfill
    \begin{subfigure}[t]{0.32\linewidth}
        \centering
        \includegraphics[width=\linewidth]{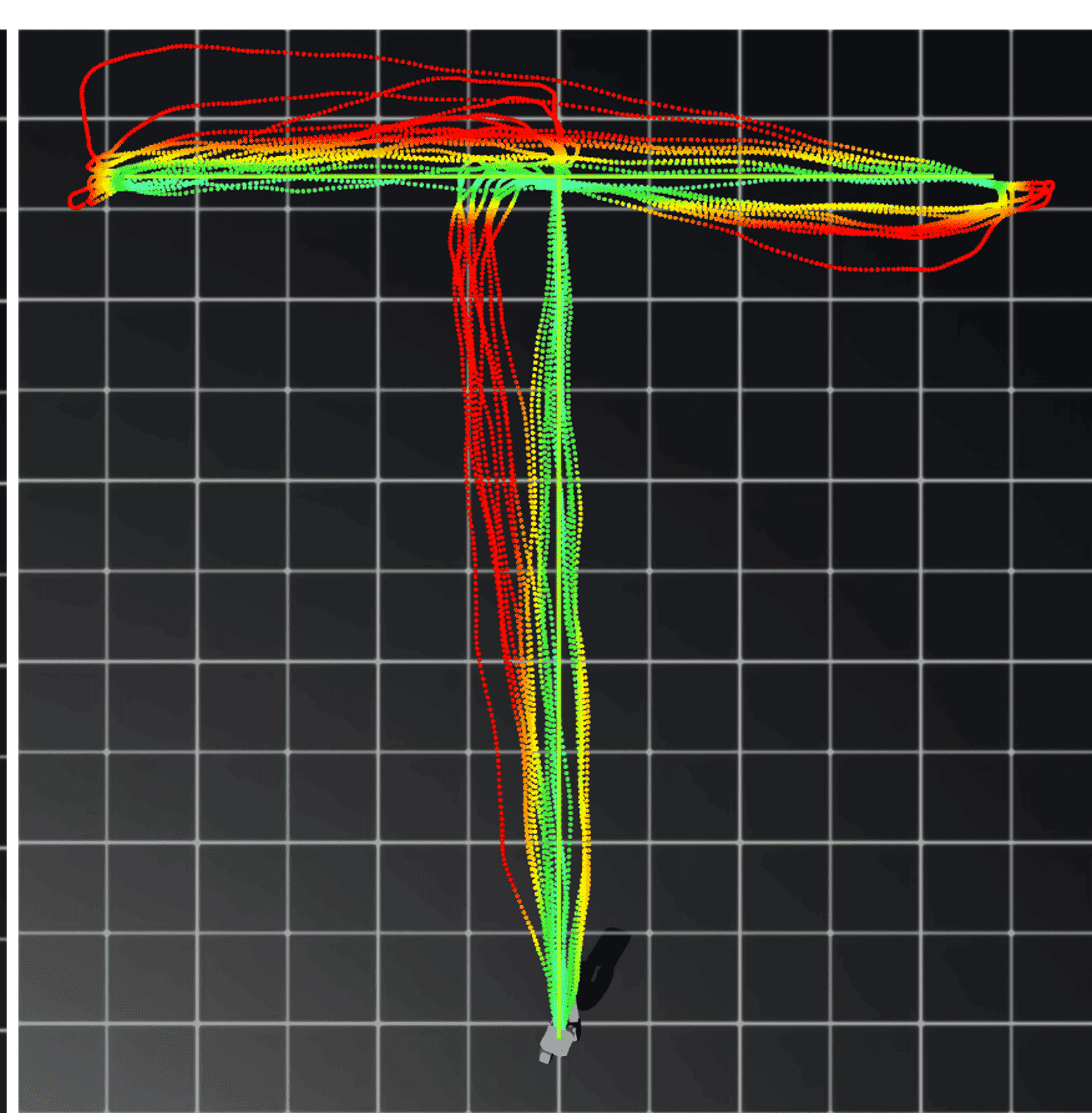}
        \subcaption{Toe-ablation}\label{fig:ttest_ablation}
    \end{subfigure}
    \caption{T-test agility comparison.}
    \label{fig:ttest}
\end{figure}

To assess how active toes affect bipedal agility, we propose a Robot T-Test inspired by the athletic
Agility T-Test [28]. Unlike the human test, the robot may rotate its torso freely and is trained with
position-command rewards rather than walking velocity-command rewards. Waypoints are issued sequentially as in
Fig.~\ref{fig:ttest}(a) along A$\rightarrow$B$\rightarrow$C$\rightarrow$D$\rightarrow$B$\rightarrow$A.

Table~\ref{tab:agility_results} reports the ten-trial averages. Completion time is nearly unchanged
(16.06~s vs.\ 16.08~s). The toe-ablation model attains higher peak and average speeds (3.549~vs.\
3.313~m/s and 2.182~vs.\ 2.085~m/s), but tracks the course less accurately: with the active toe, average and
maximum path deviation decrease by 25.0\% and 34.0\% (0.231~vs.\ 0.308~m and 0.682~vs.\ 1.033~m).
Therefore, although the toe-equipped model is relatively slower, it exhibits greater agility in turning and
stays closer to the prescribed path, finishing marginally faster overall. Fig.~\ref{fig:ttest}(b) and (c) show
the trajectories from the T-test trials. Compared with the toe-equipped case, the toe-ablation trajectories
depart from the reference path more often. This confirms that active toes improve turning agility on the
T-test.

\begin{table}[t]
    \centering
    \caption{T-test agility experiment results}
    \label{tab:agility_results}
    \small
    \setlength{\tabcolsep}{4pt}
    \renewcommand{\arraystretch}{1.3}
    \begin{tabular}{@{}l@{\hspace{1.0em}}c@{\hspace{1.0em}}c@{\hspace{0.85em}}c@{}}
    \toprule
    \textbf{Items} & \textbf{Toe-equipped} & \textbf{Toe-ablation} & \textbf{$\Delta$ [\%]} \\
    \midrule
    \rowcolor{gray!15}
    Time to goal [s]           & 16.06 & 16.08 & $-$0.1 \\
    Max speed [m/s]            & 3.313 & 3.549 & $-$6.6 \\
    \rowcolor{gray!15}
    Avg. speed [m/s]           & 2.085 & 2.182 & $-$4.4 \\
    Avg. path deviation [m]    & 0.231 & 0.308 & $-$25.0 \\
    \rowcolor{gray!15}
    Max path deviation [m]     & 0.682 & 1.033 & $-$34.0 \\
    \bottomrule
    \end{tabular}
\end{table}

\section{Conclusion}
This paper presented a 14-DOF bipedal robot with active toes and cooperative actuation, demonstrating that the active toe mechanism significantly improves energy efficiency, impact absorption, and turning agility through high-fidelity RL simulations. Although these simulation-first insights await physical hardware validation and cross-platform generalization, they provide a rigorous foundation for closing the sim-to-real gap. Future work will focus on hardware deployment, real-world locomotion testing across diverse terrains, and refining the design to maximize the practical benefits of robotic toes.

\section*{Acknowledgment}
This research was supported by the Future Mobility Project of WIRobotics Inc.

%

%

\end{document}